\def\year{2022}\relax
\documentclass[letterpaper]{article} 
\usepackage{aaai22}  
\usepackage{times}  
\usepackage{helvet}  
\usepackage{courier}  
\usepackage[hyphens]{url}  

\usepackage{graphicx} 
\usepackage{subfigure}

\usepackage{natbib}  
\usepackage{caption} 
\DeclareCaptionStyle{ruled}{labelfont=normalfont,labelsep=colon,strut=off} 
\usepackage{algorithm}
\usepackage{algorithmic}
\usepackage{amsmath}
\usepackage{amssymb}

\usepackage{newfloat}
\usepackage{listings}
\floatstyle{ruled}
\newfloat{listing}{tb}{lst}{}
\floatname{listing}{Listing}
\title{A Perceptual Distortion Reduction Framework: Towards Generating Adversarial Examples with High Perceptual Quality and Attack Success Rate}
\author {
    Ruijie Yang,\textsuperscript{\rm 1}
    Yunhong Wang, \textsuperscript{\rm 1}
    Ruikui Wang, \textsuperscript{\rm 1}
    Yuanfang Guo \textsuperscript{\rm 1}
}
\affiliations {
    \textsuperscript{\rm 1} School of Computer Science and Engineering, Beihang University, China
}

\usepackage{bibentry}

\begin{document}

\maketitle

\begin{abstract}
Most of the adversarial attack methods suffer from large perceptual distortions such as visible artifacts, when the attack strength is relatively high. These perceptual distortions contain a certain portion which contributes less to the attack success rate. This portion of distortions, which is induced by unnecessary modifications and lack of proper perceptual distortion constraint, is the target of the proposed framework. In this paper, we propose a perceptual distortion reduction framework to tackle this problem from two perspectives. Firstly, we propose a perceptual distortion constraint and add it into the objective function to jointly optimize the perceptual distortions and attack success rate. Secondly, we propose an adaptive penalty factor $\lambda$ to balance the discrepancies between different samples. Since SGD and Momentum-SGD cannot optimize our complex non-convex problem, we exploit Adam in optimization. Extensive experiments have verified the superiority of our proposed framework.
\end{abstract}

\begin{figure*}
  \centering
  \subfigure[Original Image]{
  \begin{minipage}[t]{0.14\linewidth}
    \centering
    \includegraphics[width=1in]{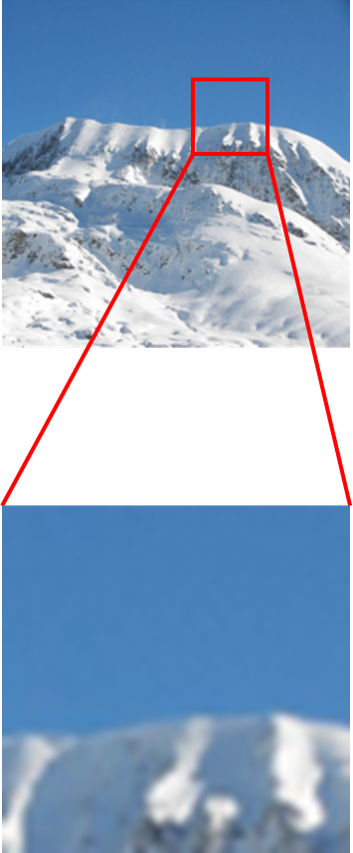}
  \end{minipage}
  }
  \subfigure[FGSM]{
  \begin{minipage}[t]{0.14\linewidth}
    \centering
    \includegraphics[width=1in]{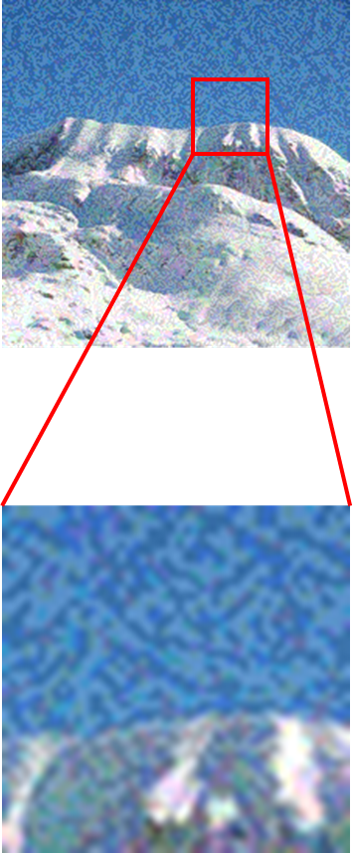}
  \end{minipage}
  }
    \subfigure[ILA]{
  \begin{minipage}[t]{0.14\linewidth}
    \centering
    \includegraphics[width=1in]{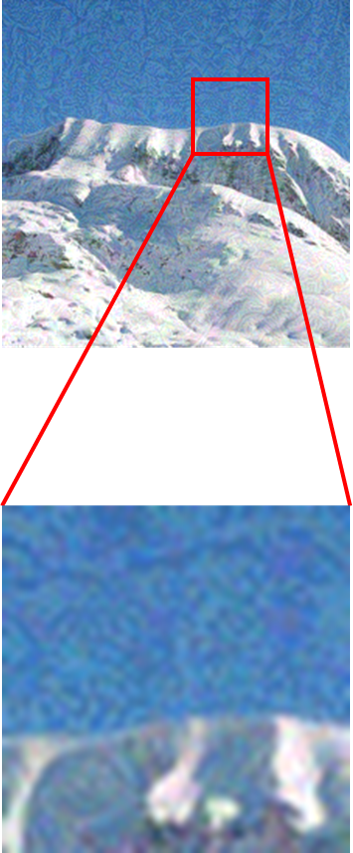}
  \end{minipage}
  }
    \subfigure[DIM]{
  \begin{minipage}[t]{0.14\linewidth}
    \centering
    \includegraphics[width=1in]{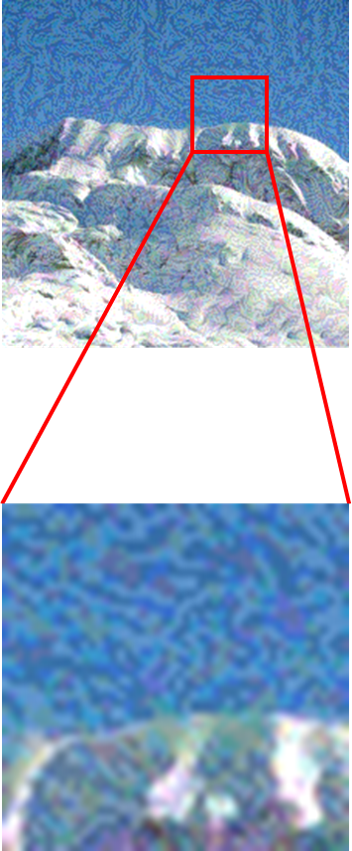}
  \end{minipage}
  }
    \subfigure[TI-DIM]{
  \begin{minipage}[t]{0.14\linewidth}
    \centering
    \includegraphics[width=1in]{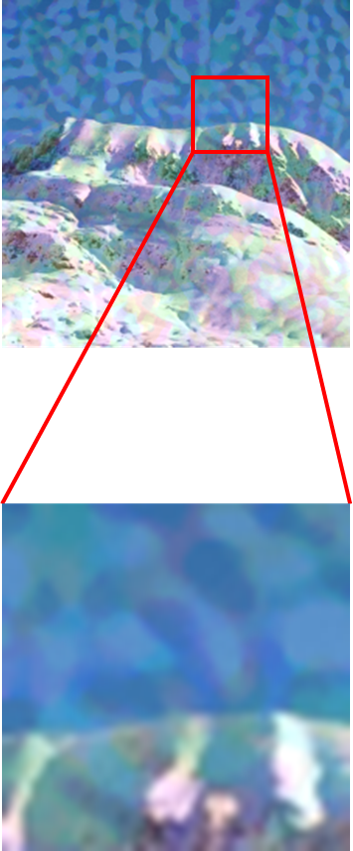}
  \end{minipage}
  }
    \subfigure[DIM-PDR(Ours)]{
  \begin{minipage}[t]{0.14\linewidth}
    \centering
    \includegraphics[width=1in]{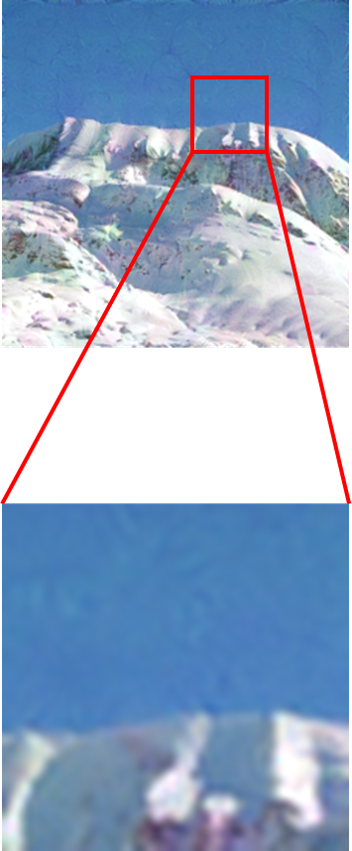}
  \end{minipage}
  }
  \caption{The adversarial examples. (a) is the original image and (b)$\sim$(f) are the adversarial examples generated by different methods. The original image (a) is correctly classified by ResNet-50. (b)$\sim$(f) are misclassified. From the generated adversarial examples, we can observe that the popular adversarial attack methods tend to induce obvious visible artifacts, regardless of white-box or black-box attacks. On the contrary, by applying our framework to a baseline adversarial attack method, the improved method (DIM-PDR) can generate adversarial examples which are visually pleasing.}
  \label{comparison}
\end{figure*}

\begin{figure}[h]
  \centering
  \includegraphics[width=0.85\linewidth]{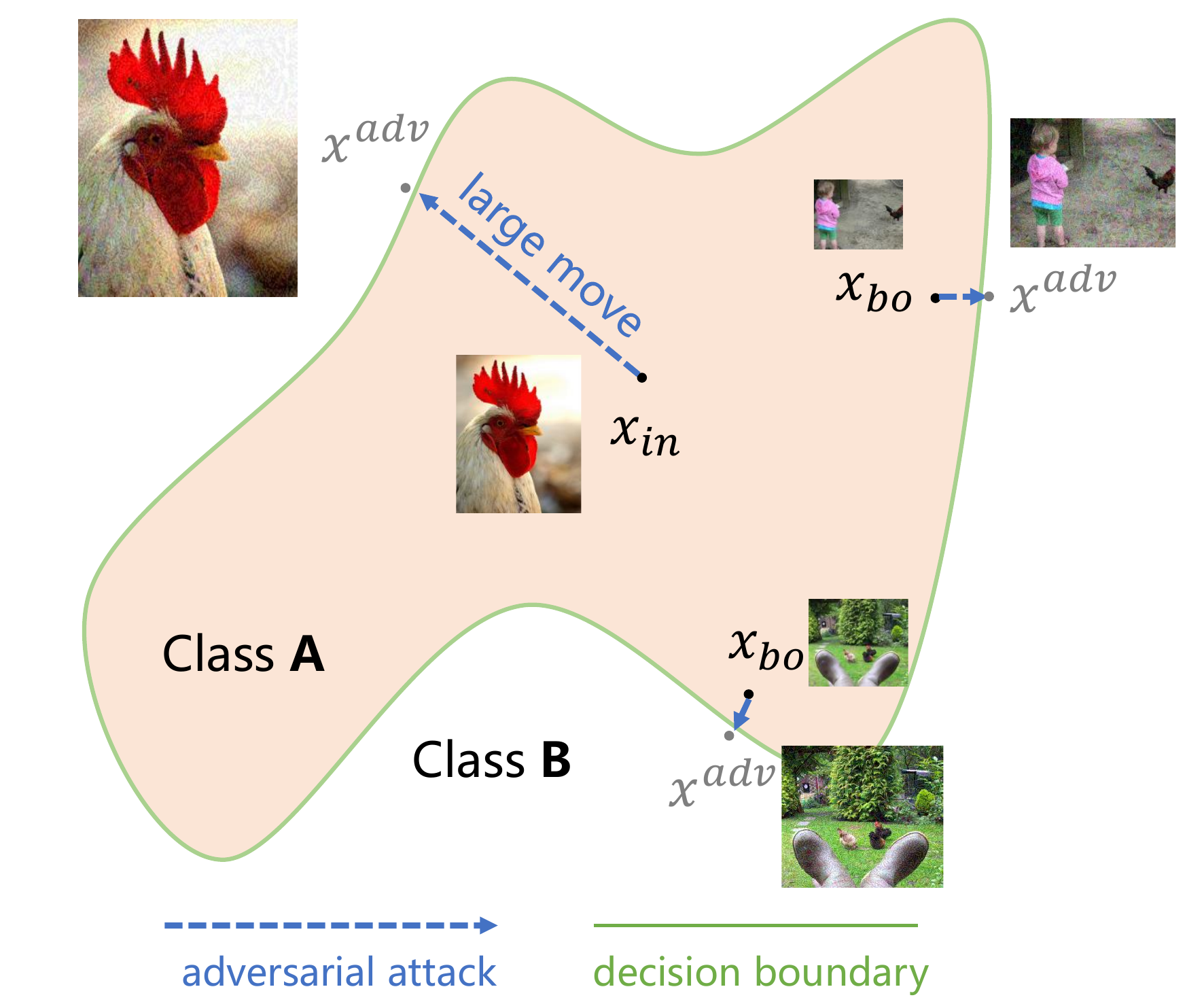}
  \caption{Schematic illustration of the differences among the samples. Some samples are located close to the decision boundary of two classes, which are represented as $x_{bo}$. Little perturbations are already enough to change their classification results. On the other hand, attacking the samples which are distant from the decision boundaries will be more difficult. Larger perturbations are required to guarantee successful attacks.}
  \label{inner-boundary}
\end{figure}

\section{Introduction}
Deep Neural Networks (DNNs) have achieved tremendous success in various vision tasks, such as image classification, object detection, etc. Unfortunately, these DNNs, which can be easily fooled by a special kind of small perturbations to give false outputs, are surprisingly vulnerable~\cite{Heaven2019WhyDA}. This type of small perturbations, named adversarial perturbations~\cite{szegedy2013intriguing}, is usually specifically generated by the adversary and added to the inputs of the DNNs to generate certain modified inputs, i.e., the adversarial examples, to attack the DNNs. Numerous researches~\cite{goodfellow2014explaining,kurakin2016adversarial,madry2017towards,carlini2017towards,zheng2019distributionally} have been subsequently conducted on the adversarial attacks to investigate and understand the vulnerabilities and the mechanisms of DNNs. Then, researchers can develop DNNs with more robustness and precise explanations.

Typical adversarial examples, which are generated by typical adversarial attack techniques, tend to possess obvious perceptual distortions, as shown in Fig.~\ref{comparison}, when a high attack success rate (ASR) is demanded. These obvious perceptual distortions are easily to be spotted by human perceptions, which prevents the adversarial examples from being eliminated. We believe that a portion of these perceptual distortions are mainly induced by unnecessary modifications and lack of proper constraint.

The existing adversarial attack methods usually equally treat all the pixels in the input image. However, different pixels in an image tend to contribute differently to the output of the DNNs~\cite{selvaraju2020grad}. Therefore, a more accurate modeling should take this into consideration. The pixels, which are important to the classification results, are desired to be applied with more perturbations, and vice versa.

Most of the existing adversarial attack methods assume the modifications on inputs are constrained by a $L_p$ ball. Typically, it can be formulated as $||x^{adv}-x||_p<\epsilon$, where $x^{adv}$ and $x$ represent the perturbed and original inputs, respectively, and $\epsilon$ is a predefined small constant which controls the maximum distance between $x^{adv}$ and $x$. As mentioned in~\cite{zhang2011fsim}, the $L_p$ ball constraint is inconsistent with human visual perception. Consequently, the $L_p$ ball constraint may induce visible artifacts, as shown in Fig.~\ref{comparison}.

Currently, there exists several literatures~\cite{superpixel,gopfert2020adversarial,deng2020frequency,perceptionimprovement}, which alleviate the distortions from different perspectives.~\cite{superpixel} focuses on adding the perturbations only to the activated regions of the target model, which ignores the potentials of the other pixels. Meanwhile,~\cite{gopfert2020adversarial,deng2020frequency,perceptionimprovement} only constrain the perturbations by adaptively thresholding them in spatial/frequency domains, according to certain criteria such as just noticeable difference (JND), etc. They, as well as the other adversarial attack methods, have not explicitly considered the tradeoff between the attack success rate and the perceptual distortions.

In this paper, we focus on reducing the perceptual distortions in the adversarial examples while maintaining a high attack success rate. By explicitly considering the trade-off between the attacking ability and the perceptual distortions, we propose a perceptual distortion reduction (PDR) framework for adversarial perturbation generation, to reduce the visible artifacts while maintaining a high attack success rate. Specifically, we propose a perceptual distortion regularization term in the final objective function with a factor $\lambda$ to model the tradeoff between the attack success rate and perceptual distortions. Meanwhile, as can be observed from Fig~\ref{inner-boundary}, there exist certain samples, which are close to the decision boundary and can be easily moved across the decision boundary with little movements(perturbations). On the contrary, other samples tend to require large movements to be moved across the boundary. We propose to adaptively calculate the factor $\lambda$ to subtly model the differences among samples. By applying the proposed framework to the existing iterative attack methods, the modified methods can still preserve decent attack success rates, with obvious perceptual distortion reductions. Since the proposed optimization problem is a more complex non-convex problem compared to the existing objective functions, we exploit Adam~\cite{Adam}, as a our optimizer to effectively solve our problem. Empirical results also verify that Adam can converge much better and faster for our framework. Note that our framework can be applied to various tasks, such as object detection ,segmentation, etc. For demonstration, we employ image classification task as an example in this paper.

Our contributions are summarized as follows:
\begin{itemize}
  \item We propose a perceptual distortion reduction (PDR) framework, which can be applied to the existing iterative adversarial perturbation generation methods, to effectively reduce the perceptual distortions while maintaining a high attack success rate.
  \item We propose to add a perceptual distortion constraint into the objective function, which can explicitly model the tradeoff between the attack success rate and perceptual distortions.
  \item We propose a new approach to adaptively calculate the factor $\lambda$, to precisely balance the discrepancies between different samples.
  \item We exploit Adam as the optimizer to effectively solve the complex non-convex optimization problem and achieve a fast convergence.
\end{itemize}

\section{Preliminaries and Related Work}

\subsection{Notations}
In this paper, we represent a K-class Deep Neural Network (DNN) classifier as $f: [0, 1]^{C \times W \times H} = X \rightarrow Y = R^K$, where $X$ stands for the input space with dimension $Channel \times Width\times Height$ and Y denotes the classification space with K categories. Let $x$ be the original input and $y$ be the corresponding true label. The output of the classifier is represented as $f(x)=(c_1,c_2,...,c_i,...,c_K)$, where $c_i$ denotes the confidence score for class $i$. The adversarial counterpart of $x$ is written as $x^{adv} = x + \delta$. This adversarial counterpart can be generated by an attack method $F$, i.e., $x^{adv} = F(f, x, y)$. Note that $x^{adv}$ is usually desired to be visually similar to $x$. Most of the attack methods constrain $||\delta||_p \le \epsilon$ by a predefined small constant $\epsilon$. The norm $||\cdot||_p$ usually employs $p=\infty$, $p=2$ or $p=0$. For instance, if $p=\infty$, the constant $\epsilon$ is usually set to $16/255$ or $8/255$.

\subsection{Adversarial Attack}

Here, we introduce the milestone methods for generating adversarial examples.

\noindent \textbf{Fast Gradient Sign Method(FGSM).}  FGSM~\cite{goodfellow2014explaining} is proposed as an one-step method, which adds a maximum perturbation to the inputs in the direction that increases the classification loss $J(x, y)$:
\begin{equation}\label{fgsm}
  x^{adv} = x + \epsilon \cdot sign(\nabla_xJ(x, y)).
\end{equation}

\noindent \textbf{Iterative Fast Gradient Sign Method(I-FGSM).} I-FGSM~\cite{kurakin2016adversarial} extends FGSM by iteratively adding multiple small perturbations and adjusting the direction of the perturbation addition in each iteration, as Eq.~\ref{i-fgsm} shows.
\begin{equation}\label{i-fgsm}
 \begin{split}
    x_0^{adv} & = x \\
    x_t^{adv} = Clip_{x, \epsilon} \{x_{t-1}^{adv}  & + \alpha \cdot sign(\nabla_{x} J(x_{t-1}^{adv}, y))\}
 \end{split}
\end{equation}
where $Clip_{x, \epsilon}\{\cdot\}$ constrains the adversarial example $x_t^{adv}$ inside an $\epsilon$-ball around $x$, and $\alpha$ denotes the step size in each iteration.

\noindent \textbf{Momentum Iterative Fast Gradient Sign Method(MI-FGSM).} MI-FGSM~\cite{dong2018boosting} introduces the momentum mechanism, which accumulates the history optimization directions to guide the current direction, into the process of generating adversarial examples. It is formulated as
\begin{equation}\label{mi-fgsm}
 \begin{split}
    g_t &= m \cdot g_{t-1}  + \frac{\nabla_{x} J(x_{t-1}^{adv}, y)}{||\nabla_{x} J(x_{t-1}^{adv}, y)||_1}, \\
    x_t^{adv} &= Cilp_{x, \epsilon}  (x_{t-1}^{adv} + \alpha \cdot sign(g_t)).
 \end{split}
\end{equation}

\noindent \textbf{Diversity Input Method(DIM).} Instead of modifying the direction in each iteration, DIM~\cite{DIM} utilizes the input diversity to improve the performance of adversarial examples. Before an image is sent into the network $C$, a random transformation $T$ is applied to the image $x$ with a transformation probability $p \in [0, 1]$, as
\begin{equation}\label{dim}
  \begin{split}
     g_t  &= \nabla_x J(T (x_{t-1}^{adv}, p), y), \\
     x_t^{adv} &=  Clip_{x, \epsilon}(x_{t-1}^{adv} + \alpha \cdot sign(g_t)).
  \end{split}
\end{equation}

\noindent \textbf{Translation-Invariant attack Method(TIM).} TIM~\cite{TIM} proposes to use a series of translated images to jointly optimize the adversarial perturbation, which improves the transferability. Besides, they approximate this process by a pre-defined kernel $W$. Therefore, the updating rule of TIM possesses the following form.
\begin{equation}\label{tim}
  x_t^{adv} = x_{t-1}^{adv} + \alpha \cdot sign(W * \nabla_x J(x_{t-1}^{adv},y))
\end{equation}
TIM can be further integrated into DIM to construct TI-DIM, which can be formulated as
\begin{equation}\label{tim}
  x_t^{adv} = x_{t-1}^{adv} + \alpha \cdot sign(W * \nabla_x J(T (x_{t-1}^{adv}, p),y)).
\end{equation}

\noindent \textbf{Intermediate-Level Attack(ILA).} Different from other attack methods, ILA~\cite{ILA} injects its perturbations onto a pre-specified intermediate layer of the source model to achieve a better black-box transferability. It employs an existing adversarial example $x^{ref}$ as the initial state. Then, two intermediate level losses are proposed. The Intermediate Level Attack Projection (ILAP) loss is defined as
\begin{equation}\label{ILAP}
  L(x^{adv}) = -(f_l(x^{ref})-f_l(x)) \cdot (f_l(x^{adv})-f_l(x)),
\end{equation}
where $f_l()$ represents the output of the pre-specified layer $l$.
The Intermediate Level Attack Flexible(ILAF) loss is formulated as
\begin{equation}\label{ILAF}
 \begin{split}
      L(x^{adv}) = &- C \cdot \frac{||f_l(x^{adv})-f_l(x)||_2}{||f_l(x^{ref})-f_l(x)||_2} \\
  &- \frac{f_l(x^{adv})-f_l(x)}{||f_l(x^{adv})-f_l(x)||_2} \cdot \frac{f_l(x^{ref})-f_l(x)}{||f_l(x^{ref})-f_l(x)||_2},
 \end{split}
\end{equation}
where $C$ is a constant which controls the balance between these two terms.
Based on ILA, ILA++~\cite{ila-plus} not only employs the existing adversarial example as the initial state, but also makes use of the intermediate results in the baseline phase of generating adversarial examples.

\subsection{Perceptual Distortion Evaluation}
In adversarial attacks, the basic image distortion measurement is the $L_p$ distance, which is easy to calculate and convenient to apply. However, this distortion measure, which is similar to the conventional metrics such as peak signal-to-noise ratio (PSNR) and mean-squared error (MSE), does not match human visual system (HVS) well. In the past two decades, a large number of image quality assessment metrics, which were designed based on human perception mechanisms, have been proposed. The famous structure similarity (SSIM) index~\cite{wang2004image} is developed based on the hypothesis that HVS is highly sensitive to the structural information in images. Subsequently, feature similarity (FSIM) index~\cite{zhang2011fsim} assumes that phase congruency is the primary feature in HVS. Recently, DNNs have also been adopted to evaluate the perceptual distortions in images~\cite{zhang2018perceptual,DeepVirtualRealityImageQuality}.

There exist many perceptual distortion evaluation methods which can behave decently in simulating human visual system. We can adopt them to evaluate the perceptual distortions induced by the generated adversarial perturbations and produce adversarial examples with better imperceptibility.

\section{Proposed Work}

\subsection{Overall Framework}
As illustrated in the above section, the process of generating adversarial examples can be considered as an optimization problem. From this perspective, a typical adversarial attack method usually contains an objective function (e.g. a Cross Entropy loss function), a feasible region $R$, (such as, $R = \{x^{adv}, ||x^{adv}-x||_{\infty}<\epsilon\}$) and optimization algorithms. Most of the existing methods modify one or more aspects. Carlini \& Wagner\'s method~\cite{carlini2017towards} and ILA~\cite{ILA} improve the objective function. I-FGSM~\cite{kurakin2016adversarial}, MI-FGSM~\cite{dong2018boosting} and DIM~\cite{DIM} introduce new optimization algorithms. These methods usually focus on achieving a high attack success rate. Unfortunately, they also generate visible artifacts in the adversarial examples, as shown in Fig.~\ref{comparison}. According to our understanding, the perceptual distortions in typical adversarial examples are mainly induced by the lack of proper constraint for perceptual distortions.

To effectively reduce the perceptual distortions while maintaining a high attack success rate, we propose a novel framework, named perceptual distortion reduction (PDR) framework, by improving the objective function and the optimization algorithm. In the improved objective function, we explicitly model the tradeoff between the perceptual distortions and attack success rate by adding a perceptual distortion regularization term. Then, the perceptual distortions and attack success rate can be jointly optimized. Since different input samples tend to possess different distance to the decision boundaries, we leverage an adaptive factor $\lambda$ in the final optimization function to balance the discrepancies between different samples. Since our formulation introduce a more complex non-convex problem, we exploit Adam as our optimizer, whose validity and effectiveness have been empirically demonstrated in the latter section. Note that the process of our method is summarized in Algorithm~\ref{alg:algorithm}.

\subsection{Perceptual Distortion Constraint}
The existing adversarial attack methods usually constrain the perturbation within a $L_p$ ball by clipping the excessive perturbations to the maximal allowed value. Unfortunately, the $L_p$ distance is inconsistent with human visual system~\cite{wang2004image, zhang2011fsim}, and the generated adversarial examples tend to contain obvious visible artifacts. To reduce these artifacts, a straightforward approach is to add an perceptual distortion constraint $L_{PD}$ to the final objective function of the adversarial perturbation generation, to jointly optimize the attack success rate and the visual quality for each example. However, this constraint may not be suitable to be directly added to the objective function, because the maximum allowed amplitude of the distortions is usually inversely proportional to the attack success rate. Therefore, we control the tradeoff between the perceptual distortions and the attack strength via a factor $\lambda$. By maintaining the optimization term for misclassification $L_{mis}$, the objective function can be formed as
\begin{equation}\label{L-total}
  L_{total}(x^{adv};x,y,f) = L_{mis}(x^{adv},y,f)+\lambda L_{PD}(x,x^{adv}).
\end{equation}
Note that $L_{mis}$ will employ the objective function of specific adversarial attack method, to which the proposed framework is applied. $L_{PD}$ can exploit any reasonable perceptual distortions models. To demonstrate the effectiveness of the proposed perceptual distortion constraint, in this paper, we utilize the popular metric $SSIM$ as $L_{PD}$.

\begin{algorithm}[tb]
\caption{The PDR Framework}
\label{alg:algorithm}
\textbf{Input}: image $x$, label $y$, target network $f(\cdot)$, adversarial objective function $L_{mis}$, perceptual distortion constraint $L_{PD}$\\
\textbf{Parameter}: $\lambda$, $lr$, $T$, $\epsilon$\\
\textbf{Output}: adversarial example $x^{adv}$
\begin{algorithmic}[1] 
\STATE $x^{adv}_0 = x$
\STATE $\lambda_0 = \lambda$
\STATE Let $k=0$.
\STATE Use $L_{total}= - L_{mis} - \lambda_k (L_{PD}-T)$ to replace the objective function.
\WHILE{True}
\STATE $k=k+1$
\STATE $g_k = \nabla_x L_{total}$
\STATE $x^{adv}_k=Adam(x^{adv}_{k-1}, g_k; \alpha, \beta_1, \beta_2, \epsilon_a)$ (use Adam optimization method to find the minima)
\STATE $\lambda_k = \lambda_{k-1} + lr \cdot \frac{\partial L_{total}}{\partial \lambda_{k}}$  
\STATE $x^{adv}_k=Clip_{x, \epsilon}(x^{adv}_k)$
\IF {$x^{adv}_k$ satisfies the termination condition.}
\STATE Break
\ENDIF
\ENDWHILE

\STATE \textbf{return} $x^{adv}$=$x^{adv}_k$
\end{algorithmic}
\end{algorithm}

\subsection{Rethinking of the Objective Function}
Suppose we have an optimization problem $\min \ \textnormal{-}L_{mis}$, which equals to $\max \ L_{mis}$, with feasible region $L_{PD}(x^{adv}, x) \ge T$, which is formulated as
\begin{equation}\label{new-perspective}
  \begin{split}
       & minimize \quad -L_{mis}(x^{adv}, y, f)\\
       & subject \ to \quad L_{PD}(x^{adv}, x) \ge T.
  \end{split}
\end{equation}
Here, we assume that a higher value of $L_{PD}$ represents a higher visual quality (if not, $\textnormal{-}L_{PD}$ can be used to replace $L_{PD}$). Typically, the feasible region is complex non-convex and it is extremely difficult to handle. In this paper, we bypass the direct optimization of this region. Instead, we propose to add a penalty term $\max(T - L_{PD}(x^{adv}, x), 0)$ to the objective function as below, with a penalty factor $\lambda$ which penalizes the samples outside the feasible region.
\begin{equation}\label{infty-form}
  minimize \quad -L_{mis}+\lambda \cdot \max(T - L_{PD}(x^{adv}, x), 0)
\end{equation}
When the sample $x^{adv}$ is in the feasible region, the second term will become $0$. If the sample $x^{adv}$ is outside the feasible region, i.e., $L_{PD}(x^{adv}, x) < T$, the penalty term will play an important role in retaining the perceptual quality. When the penalty factor $\lambda$ increases, the penalty term tends to have great effects to the objective function, if $x^{adv}$ is outside the feasible region. When $\lambda \rightarrow +\infty$, the penalty term will force $x^{adv}$ to move into the feasible region. Otherwise, $L_{total}$ will approach $\infty$. Unfortunately, we cannot directly utilize Eq.~\ref{infty-form} to generate adversarial examples because of the following reasons. Firstly, as mentioned above, $\lambda$ is required to approach $\infty$ to guarantee the convergence, which tends to introduce extra hyper-parameters such as the incremental step of $\lambda$ and the specific approximation of $\infty$. Apparently, this iterative $\infty$ approaching scheme tends to induce excessive computations due to the convergence requirement. Therefore, Eq.~\ref{infty-form} can hardly be applied in real scenarios. Besides, this model ignores the discrepancies between different samples, which will be specifically explained later. Therefore, we improve Eq.~\ref{infty-form} to solve the above deficits in next subsection.

\subsection{Adaptively Adjusted Factor $\lambda$}
Since some samples may be located close to a classification boundary while others are located relatively far from the boundaries, as shown in Fig.~\ref{inner-boundary}, different samples tend to possess diverse attack difficulties. Therefore, a fixed constant $\lambda$ is inappropriate for the diverse samples. When a sample is located close to the classification boundary, it is easier to make successful attack and induces a higher value in $L_{mis}$. Under this circumstance, the weight of the perceptual distortion constraint can be increased to give a better perceptual quality, i.e., a larger $\lambda$ is more suitable, because modifications with small amplitude can already achieve the attack. On the contrary, when a sample is located relatively far from the decision boundaries, it is more difficult to move it to another class. Then, it will cause a relatively smaller value in $L_{mis}$. In this situation, to make a successful attack, larger perturbations are necessary to be added to the samples, i.e., a small $\lambda$ tends to be more proper. Besides, using $\max(\cdot)$ function to constrain the penalty term $\ge 0$ as that in Eq.~\ref{infty-form} is less effective, because it actually prevents the model to seek for adversarial examples with higher visual quality. Without the $\max(\cdot)$ function, if $T - ssim(x^{adv}, x) < 0$, the value of $L_{total}$ will increase as well.

By considering that a large $\lambda$ is desired with a high $L_{mis}$ and vice versa, as well as the excessive computations in optimizing Eq.~\ref{infty-form}, we relax Eq.~\ref{infty-form} and further leverage an adaptive factor $\lambda$, and obtain a relaxed objective function, as shown in Eq.~\ref{adaptive-lambda}.
\begin{equation}\label{adaptive-lambda}
  \begin{split}
    maximize \quad   L_{total} = & L_{mis} +\lambda_{k} (L_{PD}(x,x_{adv}) - T) \\
    \lambda_{k+1} = & \lambda_{k} - lr \cdot \frac{\partial L_{total}}{\partial \lambda_{k}}
  \end{split}
\end{equation}
where $lr$ is the learning rate of $\lambda$ and $T$ is a threshold to control the perceptual quality, similar to that in Eq.~\ref{new-perspective}. A higher threshold will give higher perceptual quality while losing certain attack success rate, and vice versa. For example, when an adversarial example causes little perceptual distortion ($L_{PD} > T$), this example usually possesses a weak attack strength. The penalty factor $\lambda$ will decline, so $L_{total}$ will rely on $L_{mis}$ more. In next iteration, our method tends to seek for a larger perturbation to increase the attack strength. On the contrary, when an example possesses very large distortion($L_{PD} < T$), the penalty factor $\lambda$ will increase. Then, our method tends to generate an adversarial example with less perceptual distortion. In general, the automatically calculated $\lambda$ will adaptively balance the attack strength and perceptual distortion among different images, which may be close to or distant from the decision boundaries.

\subsection{Optimization Algorithm}
Since the proposed optimization problem is a more complex non-convex problem compared to the existing objective functions, the existing SGD as well as the Momentum-SGD can hardly achieve convergence, according to our experiments, we exploit Adaptive momentum estimation (Adam)~\cite{Adam} as our optimizer to solve our optimization problem.

According to our experiments, Adam converges better and faster with our proposed formulation. When $SSIM$ is employed as $L_{PD}$ as a guidance, Adam converges in similar time to the existing methods. Since Eq.~\ref{adaptive-lambda} still seek for the maxima of the objective function, while Adam is designed to seek for the minima of the objective function, we replace $L_{total}$ in Eq.~\ref{adaptive-lambda} with $\textnormal{-}L_{total}$ in our implementations. Then, Adam can be directly applied to solve our optimization problem.

\begin{table*}[t]
\centering
\begin{tabular}{lccccc}
\hline
\hline
Attack Method &             MobileNet-v2  & DenseNet-121 & ResNet-101      & SSIM    & LPIPS \\
\hline
\hline
MI-FGSM($\epsilon$=16/255)            & 92.7     & 93.3  & 92.2      &0.6863   &0.3760\\
MI-FGSM($\epsilon$=8/255)             & 80.6     & 81.1  & 80.2      &0.8715   &0.2547\\
MI-FGSM($\epsilon$=4/255)             & 60.4     & 57.4  & 58.5      &0.9580   &0.1507\\
MI-FGSM($\epsilon$=2/255)             & 38.7     & 37.2  & 32.3      &0.9870   &0.0671\\
\hline
MI-FGSM-PDR($T=0.92$)             & 91.5     & 92.3  &92.5       &0.9176   &0.3327\\
MI-FGSM-PDR($T=0.96$)             & 85.4     & 87.2  &90.1       &0.9575   &0.2756\\
MI-FGSM-PDR($T=0.98$)            & 80.2     & 81.8  &83.7       &0.9741   &0.2284\\
MI-FGSM-PDR($T=0.999$)           & 72.9     & 74.9  &76.1       &0.9840   &0.1799\\
\hline
\hline
DIM($\epsilon$=16/255)            & 93.1     & 96.6  &97.4       &0.5854   &0.4145\\
DIM($\epsilon$=8/255)             & 81.7     & 86.9  &91.1       &0.7941   &0.2653\\
DIM($\epsilon$=4/255)             & 56.9     & 62.2  &66.9       &0.9269   &0.1389\\
DIM($\epsilon$=2/255)             & 30.2     & 34.5  &29.6       &0.9787   &0.0586  \\
\hline
DIM-PDR($T=0.92$)             & 91.6     & 97.0  &97.8       &0.9185   &0.2435\\
DIM-PDR($T=0.96$)             & 83.8     & 93.6  &96.0       &0.9570   &0.1765\\
DIM-PDR($T=0.98$)            & 75.2     & 89.0  &92.3       &0.9721   &0.1338\\
DIM-PDR($T=0.999$)           & 66.8     & 79.6  &87.6       &0.9823   &0.0967\\
\hline
\hline
TIDIM($\epsilon$=16/255)            & 92.3     & 97.8   & 97.5      & 0.7276   & 0.3604\\
TIDIM($\epsilon$=8/255)            & 73.6     & 88.8   & 84.6       & 0.8759   & 0.2133\\
TIDIM($\epsilon$=4/255)             & 47.4     & 63.9   & 53.6       & 0.9556   & 0.1014\\
TIDIM($\epsilon$=2/255)             & 29.1     & 36.6   & 28.0      &0.9861   &0.0377\\
\hline
TIDIM-PDR($T=0.92$)             & 87.9     & 95.5  &94.6       &0.9212   &0.2301\\
TIDIM-PDR($T=0.96$)             & 78.2     & 93.1  &90.5       &0.9578   &0.1687\\
TIDIM-PDR($T=0.98$)             & 67.1     & 87.9  &85.5       &0.9737   &0.1189\\
TIDIM-PDR($T=0.999$)            & 55.1     & 76.5  &70.6       &0.9750   &0.0823\\
\hline
\hline
\end{tabular}
\caption{Results of the baseline methods MI-FGSM, DIM, TI-DIM, and their corresponding improved methods. Note that X-PDR is the method that applied our PDR framework, where X denotes the baseline method. The numbers in columns 2-5 represent the ASR(\%). The results of SSIM and LPIPS are obtained by computing the average result for all the samples. }
\label{compare with MI-FGSM}
\end{table*}

\begin{figure}[t]
  \centering
  \includegraphics[width=0.80\linewidth]{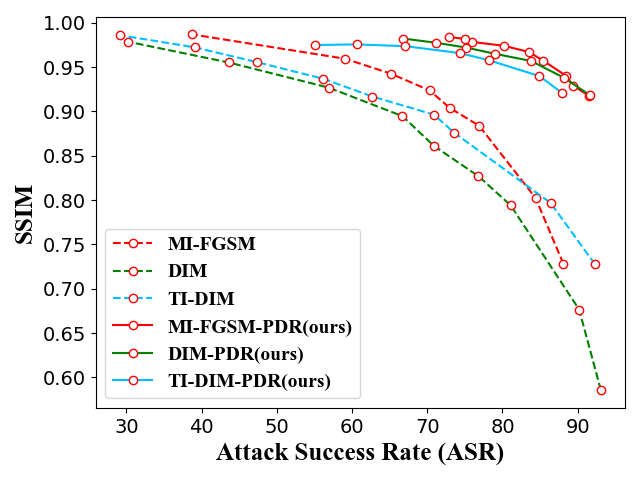}
  \caption{The curve of ASR vs. SSIM. We test all the 1000 images for the pre-trained MobileNet-v2. SSIM is obtained by computing the average result for all the samples. X-PDR represents the method applied our proposed PDR, where X denotes the baseline method.}
  \label{results}
\end{figure}


\section{Experiments}

In this section, the proposed PDR framework will be applied to several popular adversarial attack methods, where the improved method is named as X-PDR, to evaluate the effectiveness of the proposed work.

\subsection{Setups}\label{setting}
In the experiments, we follow the existing protocol~\cite{SI-NI-FGSM}  and randomly choose 1000 images from the ILSVRC 2012~\cite{imagenet} validation set, which can almost be correctly classified by all the testing models. These images are resized to $224 \times 224$ before sending to the networks. Note that our PDR framework can be applied to arbitrary iterative adversarial perturbation generation methods. However, since most white-box attacks, such as I-FGSM, PGD, AutoAttack~\cite{autoattack}, tend to generate little perturbations with a high attack success rate, they can already generate adversarial examples with decent perceptual quality. To demonstrate the effectiveness of our method, we select four popular black-box iterative methods, including MI-FGSM, DIM, TI-DIM, ILA and ILA++. Since MI-FGSM needs an ensemble model as the source model, we select the pre-trained ResNet-50~\cite{resnet}, Inception-v3~\cite{inception-v3} and VGG-16~\cite{mobilenet-v2} to construct it. For ILA, the pre-trained ResNet-18 is selected to be the source model as recommended by the authors. For the rest of the methods, the pre-trained ResNet-50 is employed to generate the adversarial examples.

For assessing the performances, we calculate the attack success rate and perceptual distortion metrics. The attack success rate (ASR) is defined as
\begin{equation}\label{asr-formula}
  ASR = 1 - \frac{\#\{correct \ samples\}}{\#\{total \ samples\}}.
\end{equation}
For the baseline methods, we manipulate the maximum allowed perturbation $\epsilon$ to generate adversarial examples with different visual quality, i.e., $||x^{adv}-x||_{\infty} \le \epsilon$. For our method, we vary the threshold $T$ in Eq.~\ref{adaptive-lambda} to generate adversarial examples with different perceptual qualities. Note that the maximum number of iterations is set to 150. The learning rate of Adam optimizer is set to 0.01. For baseline methods, the learning rate is the same as in their papers. Two popular image quality assessment metrics, SSIM~\cite{wang2004image} and LPIPS~\cite{zhang2018perceptual},  are selected to evaluate the perceptual quality of the adversarial examples.

\begin{figure*}
\centering
\begin{minipage}[t]{0.3\linewidth}
  \centering
  \includegraphics[width=1\linewidth]{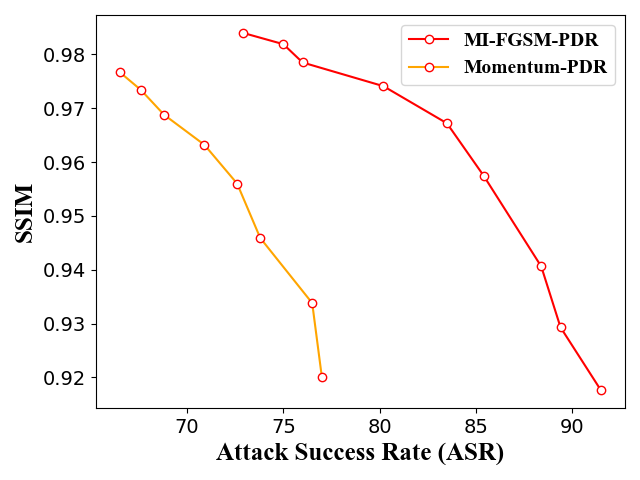}
  \caption*{(a)}
\end{minipage}
\begin{minipage}[t]{0.3\linewidth}
  \centering
  \includegraphics[width=1\linewidth]{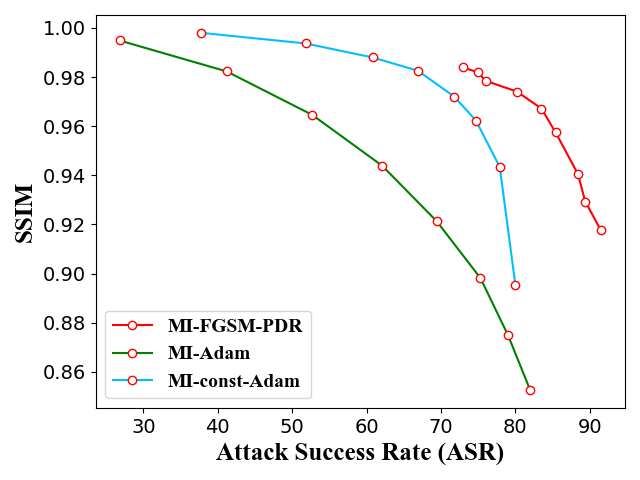}
  \caption*{(b)}
\end{minipage}
\begin{minipage}[t]{0.3\linewidth}
  \centering
  \includegraphics[width=1\linewidth]{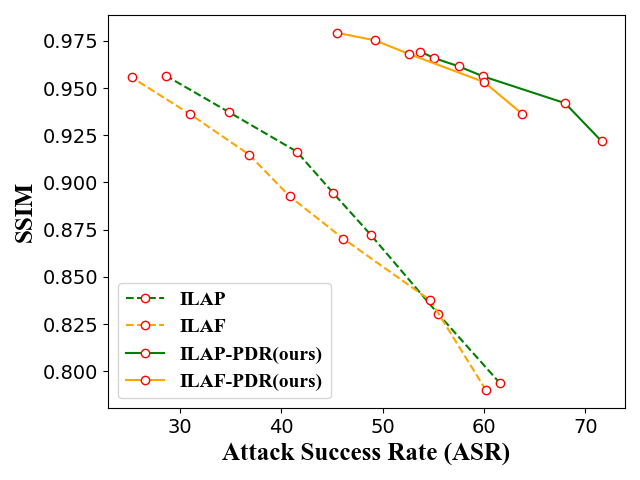}
  \caption*{(c)}
\end{minipage}
\caption{(a) is the comparison of different optimization algorithms. We compare Momentum-SGD with Adam by maintaining the other settings. (b) compares adaptive $\lambda$ with constant $\lambda$. (c) The ILA method is employed to discuss the influence of different $L_{mis}$, where we generate the adversarial examples on ResNet-18 and test them on DenseNet-121.}
\label{ablation-study}
\end{figure*}

\subsection{Results}
We compare the improved methods with their corresponding baselines. The results are shown in Table 1. In SSIM, a higher value implies a higher perceptual quality, while a smaller value implies a higher visual quality in LPIPS. The dynamic range of them are both $[0, 1]$. As can be observed, the attack success rate declines when the perceptual quality increases, for all the methods. This phenomenon verifies that there is a tradeoff between attack success rate and perceptual quality. Table 1 also reveals that the methods, which applied PDR, outperform their baseline methods by producing smaller perceptual distortions under the same ASR. For instance, if we compare row 7 with row 2, 11 and 19, the ASRs are all close to 90\% when tested on ResNet-101. Meanwhile, the corresponding SSIM value (0.957) of our method is significantly higher than the others(0.72, 0.79, 0.87).

In fact, we believe that it is unfair to just compare the attack success rate under identical $\epsilon$, because some methods actually perturbs more pixels to achieve a higher ASR, which actually yields more perceptual distortions to the images. For better illustration, we provide a figure of ASR versus SSIM, as shown in Fig.~\ref{results}. As can be obeserved, it is obvious that our method drastically improves the performance compared to the baseline methods. With the same attack strength, our method can obviously reduce the perceptual distortions and improve the perceptual quality. For example, when the attack success rate is 90\% (classification accuracy is 10\%), the SSIM value of our method is bigger than 0.90, while the others only give SSIM values lower than 0.75. If the attack success rate is less demanded, all the methods tend to generate examples with higher visual qualities. However, our method still gives the best result. Due to limited space, the figure of ASR versus LPIPS is presented in the supplementary material and it shows similar results as Fig.~\ref{results}. In summary, our method can generate adversarial examples with good perceptual qualities while preserving the attacking success rate.

\subsection{Analysis}

\noindent \textbf{Adam vs Momentum:}
To verify the influence of Adam optimization method, we conduct the following experiments. We choose MI-FGSM-PDR as the attack method and other experiment settings are set to default as previous, expect for replacing the optimization algorithm. We use Momentum-PDR represent our method with momentum optimization method. Learning rate is 0.01 and momentum is set to 0.9. Adversarial examples are all tested on pre-trained MobileNet-v2. The results are shown in Fig.~\ref{ablation-study}(a). The closer to upper right corner, the better performance. Our method has greatly improved compared to Momentum-PDR.

\noindent \textbf{Compact of adaptive $\lambda$:} Then we study the effect of adaptive $\lambda$. Here, we launch three experiments for better comparison: with adaptive $\lambda$, with constant $\lambda$ and without $\lambda$. We utilize MI-FGSM method for demonstration. With constant $\lambda$ and without $\lambda$ are represented as MI-const-Adam and MI-Adam respectively. For constant $\lambda$, we set $\lambda=400, 800, 1200, 1600, 2400, 3200, 5000, 9999$.  Other settings are set to default. The result is shown in Fig.~\ref{ablation-study}(b). We find that constant $\lambda$ get a better performance than MI-Adam, this improves the effectiveness of the perceptual distortion constraint term. But this method does not take the difference between samples into account. Hence, the performance is weaker than our method.

\noindent \textbf{Compact of different loss function $L_{mis}$:} Here we use ILA and ILA++ to study whether different loss function will affect the result of our method. Due to the limited space, the results of ILA++ are placed in the supplementary material. Instead of using CrossEntropy loss, ILA proposes two loss function: ILAP and ILAF. We can also apply our framework to this method through replacing its objective function and its optimization algorithm as described in Algorithm~\ref{alg:algorithm}. As previous, we use ILAP-PDR and ILAF-PDR to represent the methods, respectively. Following the settings in paper~\cite{ILA}, ResNet-18 is the source model to generate adversarial examples and the pre-specified immediate layer in Eq.~\ref{ILAP} and Eq.~\ref{ILAF} is layer4. All the adversarial examples are tested on pre-trained DenseNet-121. The results are shown in Fig.~\ref{ablation-study}(c). Although ILAP and ILAF loss have completely different form with CrossEntropy loss. Our method still improves them a lot. This further proves the effectiveness of our framework as our method does not rely on specified loss function. Additionally, from the figure, we can also find that the performance of ILAP and ILAF is very close.

\section{Conclusion}
In this paper, we propose a perceptual distortion reduction framework to alleviate the obvious visible artifacts in the adversarial examples. The proposed framework can be easily applied to typical iterative adversarial perturbation generation methods. We reduce the unnecessary modifications in the adversarial examples by proposing an additional perceptual distortion constraint for the objective function. To balance the discrepancies between different samples, we calculate an adaptive penalty factor $\lambda$ for different samples. To effectively solve our complex non-convex optimization problem, we utilize Adam as our optimizer. Extensive experiments have demonstrated the effectiveness of our proposed PDR framework.

\clearpage

\bibliography{aaai22}

\begin{thebibliography}{28}
\providecommand{\natexlab}[1]{#1}

\bibitem[{Carlini and Wagner(2017)}]{carlini2017towards}
Carlini, N.; and Wagner, D.~A. 2017.
\newblock Towards evaluating the robustness of neural networks.
\newblock In \emph{{IEEE} S \& P}, 39--57.

\bibitem[{Croce and Hein(2020)}]{autoattack}
Croce, F.; and Hein, M. 2020.
\newblock Reliable evaluation of adversarial robustness with an ensemble of
  diverse parameter-free attacks.
\newblock In \emph{{ICML}}, volume 119, 2206--2216.

\bibitem[{Deng et~al.(2009)Deng, Dong, Socher, Li, Li, and Li}]{imagenet}
Deng, J.; Dong, W.; Socher, R.; Li, L.; Li, K.; and Li, F. 2009.
\newblock ImageNet: {A} large-scale hierarchical image database.
\newblock In \emph{{IEEE} {CVPR}}, 248--255.

\bibitem[{Deng and Karam(2020)}]{deng2020frequency}
Deng, Y.; and Karam, L.~J. 2020.
\newblock Frequency-Tuned Universal Adversarial Attacks.
\newblock \emph{CoRR}, abs/2003.05549.

\bibitem[{Dong et~al.(2020)Dong, Han, Chen, Liu, Bian, Ma, Li, Wang, Zhang, and
  Yu}]{superpixel}
Dong, X.; Han, J.; Chen, D.; Liu, J.; Bian, H.; Ma, Z.; Li, H.; Wang, X.;
  Zhang, W.; and Yu, N. 2020.
\newblock Robust Superpixel-Guided Attentional Adversarial Attack.
\newblock In \emph{{IEEE/CVF} CVPR}, 12892--12901.

\bibitem[{Dong et~al.(2018)Dong, Liao, Pang, Su, Zhu, Hu, and
  Li}]{dong2018boosting}
Dong, Y.; Liao, F.; Pang, T.; Su, H.; Zhu, J.; Hu, X.; and Li, J. 2018.
\newblock Boosting adversarial attacks with momentum.
\newblock In \emph{{IEEE} CVPR}, 9185--9193.

\bibitem[{Dong et~al.(2019)Dong, Pang, Su, and Zhu}]{TIM}
Dong, Y.; Pang, T.; Su, H.; and Zhu, J. 2019.
\newblock Evading Defenses to Transferable Adversarial Examples by
  Translation-Invariant Attacks.
\newblock In \emph{{IEEE} {CVPR}}, 4312--4321.

\bibitem[{Goodfellow, Shlens, and Szegedy(2015)}]{goodfellow2014explaining}
Goodfellow, I.~J.; Shlens, J.; and Szegedy, C. 2015.
\newblock Explaining and harnessing adversarial examples.
\newblock In \emph{ICLR}.

\bibitem[{G{\"o}pfert et~al.(2020)G{\"o}pfert, Artelt, Wersing, and
  Hammer}]{gopfert2020adversarial}
G{\"o}pfert, J.~P.; Artelt, A.; Wersing, H.; and Hammer, B. 2020.
\newblock Adversarial attacks hidden in plain sight.
\newblock In \emph{IDA}, 235--247.

\bibitem[{He et~al.(2016)He, Zhang, Ren, and Sun}]{resnet}
He, K.; Zhang, X.; Ren, S.; and Sun, J. 2016.
\newblock Deep Residual Learning for Image Recognition.
\newblock In \emph{{IEEE} {CVPR}}, 770--778.

\bibitem[{Heaven(2019)}]{Heaven2019WhyDA}
Heaven, D. 2019.
\newblock Why deep-learning AIs are so easy to fool.
\newblock \emph{Nature}, 574: 163--166.

\bibitem[{Huang et~al.(2019)Huang, Katsman, Gu, He, Belongie, and Lim}]{ILA}
Huang, Q.; Katsman, I.; Gu, Z.; He, H.; Belongie, S.~J.; and Lim, S. 2019.
\newblock Enhancing Adversarial Example Transferability With an Intermediate
  Level Attack.
\newblock In \emph{{IEEE} {ICCV}}, 4732--4741.

\bibitem[{Kim, Lim, and Ro(2020)}]{DeepVirtualRealityImageQuality}
Kim, H.~G.; Lim, H.; and Ro, Y.~M. 2020.
\newblock Deep Virtual Reality Image Quality Assessment With Human Perception
  Guider for Omnidirectional Image.
\newblock \emph{{IEEE} TCSVT}, 30(4): 917--928.

\bibitem[{Kingma and Ba(2015)}]{Adam}
Kingma, D.~P.; and Ba, J. 2015.
\newblock Adam: {A} Method for Stochastic Optimization.
\newblock In Bengio, Y.; and LeCun, Y., eds., \emph{{ICLR} 2015}.

\bibitem[{Kurakin, Goodfellow, and Bengio(2017)}]{kurakin2016adversarial}
Kurakin, A.; Goodfellow, I.~J.; and Bengio, S. 2017.
\newblock Adversarial examples in the physical world.
\newblock In \emph{ICLR}.

\bibitem[{Li, Guo, and Chen(2020)}]{ila-plus}
Li, Q.; Guo, Y.; and Chen, H. 2020.
\newblock Yet Another Intermediate-Level Attack.
\newblock In \emph{{ECCV}}.

\bibitem[{Lin et~al.(2020)Lin, Song, He, Wang, and Hopcroft}]{SI-NI-FGSM}
Lin, J.; Song, C.; He, K.; Wang, L.; and Hopcroft, J.~E. 2020.
\newblock Nesterov Accelerated Gradient and Scale Invariance for Adversarial
  Attacks.
\newblock In \emph{{ICLR} 2020}.

\bibitem[{Madry et~al.(2018)Madry, Makelov, Schmidt, Tsipras, and
  Vladu}]{madry2017towards}
Madry, A.; Makelov, A.; Schmidt, L.; Tsipras, D.; and Vladu, A. 2018.
\newblock Towards deep learning models resistant to adversarial attacks.
\newblock In \emph{ICLR}.

\bibitem[{Sandler et~al.(2018)Sandler, Howard, Zhu, Zhmoginov, and
  Chen}]{mobilenet-v2}
Sandler, M.; Howard, A.~G.; Zhu, M.; Zhmoginov, A.; and Chen, L. 2018.
\newblock MobileNetV2: Inverted Residuals and Linear Bottlenecks.
\newblock In \emph{{IEEE} {CVPR}}, 4510--4520.

\bibitem[{Selvaraju et~al.(2020)Selvaraju, Cogswell, Das, Vedantam, Parikh, and
  Batra}]{selvaraju2020grad}
Selvaraju, R.~R.; Cogswell, M.; Das, A.; Vedantam, R.; Parikh, D.; and Batra,
  D. 2020.
\newblock Grad-CAM: Visual Explanations from Deep Networks via Gradient-Based
  Localization.
\newblock \emph{IJCV}, 128(2): 336--359.

\bibitem[{Szegedy et~al.(2016)Szegedy, Vanhoucke, Ioffe, Shlens, and
  Wojna}]{inception-v3}
Szegedy, C.; Vanhoucke, V.; Ioffe, S.; Shlens, J.; and Wojna, Z. 2016.
\newblock Rethinking the Inception Architecture for Computer Vision.
\newblock In \emph{{IEEE} {CVPR}}, 2818--2826.

\bibitem[{Szegedy et~al.(2014)Szegedy, Zaremba, Sutskever, Bruna, Erhan,
  Goodfellow, and Fergus}]{szegedy2013intriguing}
Szegedy, C.; Zaremba, W.; Sutskever, I.; Bruna, J.; Erhan, D.; Goodfellow,
  I.~J.; and Fergus, R. 2014.
\newblock Intriguing properties of neural networks.
\newblock In \emph{ICLR}.

\bibitem[{Wang et~al.(2020)Wang, Feng, Ward, Wang, and
  Wang}]{perceptionimprovement}
Wang, Y.; Feng, M.; Ward, R.; Wang, Z.~J.; and Wang, L. 2020.
\newblock Perception Improvement for Free: Exploring Imperceptible Black-box
  Adversarial Attacks on Image Classification.
\newblock \emph{CoRR}, abs/2011.05254.

\bibitem[{Wang et~al.(2004)Wang, Bovik, Sheikh, and Simoncelli}]{wang2004image}
Wang, Z.; Bovik, A.~C.; Sheikh, H.~R.; and Simoncelli, E.~P. 2004.
\newblock Image quality assessment: from error visibility to structural
  similarity.
\newblock \emph{{IEEE} TIP}, 13(4): 600--612.

\bibitem[{Xie et~al.(2019)Xie, Zhang, Zhou, Bai, Wang, Ren, and Yuille}]{DIM}
Xie, C.; Zhang, Z.; Zhou, Y.; Bai, S.; Wang, J.; Ren, Z.; and Yuille, A.~L.
  2019.
\newblock Improving Transferability of Adversarial Examples With Input
  Diversity.
\newblock In \emph{{IEEE} {CVPR}}, 2730--2739.

\bibitem[{Zhang et~al.(2011)Zhang, Zhang, Mou, and Zhang}]{zhang2011fsim}
Zhang, L.; Zhang, L.; Mou, X.; and Zhang, D. 2011.
\newblock {FSIM}: A feature similarity index for image quality assessment.
\newblock \emph{{IEEE} TIP}, 20(8): 2378--2386.

\bibitem[{Zhang et~al.(2018)Zhang, Isola, Efros, Shechtman, and
  Wang}]{zhang2018perceptual}
Zhang, R.; Isola, P.; Efros, A.~A.; Shechtman, E.; and Wang, O. 2018.
\newblock The Unreasonable Effectiveness of Deep Features as a Perceptual
  Metric.
\newblock In \emph{{IEEE} CVPR}, 586--595.

\bibitem[{Zheng, Chen, and Ren(2019)}]{zheng2019distributionally}
Zheng, T.; Chen, C.; and Ren, K. 2019.
\newblock Distributionally adversarial attack.
\newblock In \emph{AAAI}, 2253--2260.

\end{thebibliography}

\clearpage

\end{document}